# Exploring the Effects of Data Augmentation for Drivable Area Segmentation

Srinjoy Bhuiya[1*], Ayushman Kumar[2], Sankalok Sen[3]

[1]Heritage Institute of Technology, [2]TiVRA-AI, [3]The University of Hong Kong
[1]srinjoy.bhuiya.cse23@heritageit.edu.in, [2]ayushman@tivraai.com, [3]ssen2001@connect.hku.hk

**Abstract**

The real-time segmentation of drivable areas plays a vital role in accomplishing autonomous perception in cars. Recently there have been some rapid strides in the development of image segmentation models using deep learning. However, most of the advancements have been made in model architecture design. In solving any supervised deep learning problem related to segmentation, the success of the model that one builds depends upon the amount and quality of input training data we use for that model. This data should contain well-annotated varied images for better working of the segmentation model. Issues like this pertaining to annotations in a dataset can lead the model to conclude with overwhelming Type I and II errors in testing and validation, causing malicious issues when trying to tackle real world problems. To address this problem and to make our model more accurate, dynamic, and robust, data augmentation comes into usage as it helps in expanding our sample training data and making it better and more diversified overall. Hence, in our study, we focus on investigating the benefits of data augmentation by analyzing pre-existing image datasets and performing augmentations accordingly. Our results show that the performance and robustness of existing state of the art (or SOTA) models can be increased dramatically without any increase in model complexity or inference time. The augmentations decided on and used in this paper were decided only after thorough research of several other augmentation methodologies and strategies and their corresponding effects that are in widespread usage today. All our results are being reported on the widely used Cityscapes Dataset.

## 1 INTRODUCTION

In the last few years, convolutional neural networks (or CNNs) [1] have been used in every field of image processing to outperform many state of the art (SOTA) algorithms based on classical processing techniques. However, one hindrance in training DCNNs [31] has always been the availability of varied training data. Although, for our problem in relation to drivable area segmentation for autonomous driving vehicles, several open-source datasets are available which contain several thousand images for our perusal. However, a key issue in these datasets is that there exists little variance in the overall contents of these training images. For our main dataset [2], which contains five thousand images, there is not much robustness in the variance of the data. The urban environments are not only remarkably similar in nature but also there is no huge shift in weather conditions including but not limited to weather anomalies like snow, fog, heavy to light rain, and sunshine.

A couple of most critical requirements for a drivable area segmentation system such as ours to work well are precision and the ability of the system to run at real-time speed at edge computing devices (having limited computations). Keeping these above-mentioned critical requirements in mind, we have developed some strategies by which we can increase the precision of our system without increasing the overall computational requirement of our model by a huge margin.

Following Uysal et al [3], we have seen that data augmentation can be an excellent method of preventing overfitting of the model to the training data and enhancing the metrics of a CNN network implementing segmentation of various kinds. Furthermore, augmentation techniques like dropout as seen in Srivastava et al [4] have been used for controlling the rate of fitting the model on our data which enables the model to learn more complex features of the images without overfitting on the limited training images.

We have seen that the most common use of a convolutional network has been in the task of image classification where we output a single label per image sample. By using a clever encoder-decoder implementation inspired from the paper by Ronneberger et al [5], we created a lightweight fully end to end convolutional model which can generate a binary mask of the drivable area in an image with a high prediction interval, while still hypothesizing fast on a low computing strength.

To assist in the task of feature extraction and then it's respective decoding, we introduce two structures in our model namely skip connections which we take inspiration from the ResNet architecture [6] and a dual attention module inspired by Woo et al [7]. These additions result in consistent improvement in segmentation task loads with minimal overheads. **Our model reaches around 201 FPS on a light Nvidia GTX 1660ti GPU.**

In summary, our main contributions in this paper include:

- We propose an efficient encoder-decoder network architecture called CA-UNet that can handle real-time drivable area detection.
- We design ablation studies to ensure the effectiveness of our model.
- We give a detailed study of the different augmentations we can apply on our training data to improve the metrics of our model without the increase of computation requirement.

We believe this paper will provide a reference for other reliant studies in the use of augmentation and efficient model architectures in the domain of drivable area detection.

## 1.1   Related Works

Multiple advancements have been made related to our work in segmentation over the years. Late 2000s works focus on the concepts of semantic segmentation using a single monocular image, with most approaches implementing the Random Decision Forest methods. In 2008, Shotton et al [8] used Random Decision Forests on classification of local patches based "solely on motion-derived 3D world structures." Plath et al in 2009 used Conditional Random Fields (known as CRFs) [9] for fusing the local and global features of an image, implementing multi-class based image segmentation. Alongside, Sturgess et al [10] used appearance-based features and structure-from-motion (SFM) features. In 2010, Zhang et al [11] used Dense Depth Maps to perform semantic segmentation on urban sceneries. In 2011, Kontschieder et al [12] implemented class labeling techniques on Random Decision Forests for semantic image labeling techniques.

Most of these early papers used specially handcrafted features and kernels that were not sufficient for learning all aspects (both the highs and lows) of images, hindering performances of implemented models. From the early to mid-2010s, the focus was geared towards more convolutional neural network (or CNN) based architectures which performed substantially better than previously mentioned methods.

In 2012, Ciresan et al [13] introduced a sliding window approach to predict the class label of each pixel, with a surrounding patch of that pixel being fed into the model and the predictions made on those patches. The system however, had some drawbacks. Firstly, the network was to be run multiple times for a single image depending on the numbers of patches an image is divided into causing the model to be extremely slow, and secondly, the patch-based system prevents the model from learning about the global context of the image making the model have a trade-off between accuracy and context. FCN [14] revolutionized this field of study with an encoder based on VGG-16 [15], with present State of the Art (SOTA) models still inspired by this paper by Long et al in 2015. The paper introduced the first fully convolutional network to semantic segmentation workloads allowing the preservation of spatial information in the data and reducing the computational cost in comparison to fully connected layers. Using a fully convolutional neural network model frees us from the constraints of image resolutions. However, in their architecture, even after the addition of skip connections, the model failed to infer properly on images of high resolutions. In 2016, Paszke et al. [16] introduced ENet, which tried to increase the speed of model inference by reducing the size of intermediate feature maps. In 2017, SegNet [17], using an encoder-decoder architecture as well, introduced the concept of unpooling layers for upsampling, replacing transpose convolutions. It caused an increase in model training and inference times while causing a decrease in model precision overall.

In the past couple of years, several works have introduced novel improvements in overall model architectures. In 2018, Chen et al suggested a novel encoder-decoder model [18] for object boundary refinement, besides also using Atrous Spatial Pyramid Pooling (ASPP) which is an extremely helpful module regarding semantic segmentation for resampling feature-based layers multiple times prior to convolution. Then, in 2019 Tian et al [19] introduced data-dependent upsampling used for handling data redundancy in label space. Finally in 2020, the model prepared by Han et al [20] started the trend of multitask-based learning by trying to provide solutions to the tasks of edge detection and drivable area segmentation together into a single coupled network to reduce computation requirements.

## 1.2 Some Seminal Works

Despite the works presented above, some of the works worth mentioning separately which have had a massive impact in the field of image segmentation are as follows:

- **Otsu Thresholding:** [21] An algorithm used to perform image thresholding and generate binary a segmentation map automatically. The threshold is measured by amplifying inter-class variance. In its most uncomplicated way, this algorithm returns different edge data that helps separate the pixels into two foreground and background classes respectively. However, if the illuminations are non-uniform in the image, the histogram no longer stays bimodal, and thresholding becomes unsatisfactory.
- **Watershed algorithm:** [22] An image processing method, mainly used for object segmentation i.e., for separating different objects in an image. Watershed Algorithm portrays high-intensity pixels as peaks while low-intensity pixels as valleys. This algorithm has a greater advantage over traditional thresholding image processing methods because Watershed Algorithm can extract each individual detail from the image. But noisy images can influence the segmentation in the wrong direction, because at each minima, a watershed is generated.
- **U-Net:** [5] Proposes an encoder-decoder architecture. The encoder or contracting path captures context and the decoder or expanding path enables precise localization. The contracting path follows the architecture of a CNN, while the expanding path consists of up-convolutions, a concatenation with the corresponding cropped feature maps from the contracting path (skip connections) to make sure data is not lost. The authors have tried to build a network that can be trained from very few images and outperforms SOTA CNN networks at the time of authoring the paper. The only disadvantage to U-Net is the learning can slow down in deeper models, which might be a risk as it causes the neural network model to learn to discard or ignore states having abstract features.
- **Dropout:** [4] Introduces dropout as a method to prevent the model from overfitting on training split. Standard backpropagation methodologies learn well on the training split but do not generalize to unseen data. Dropout helps break this learning and helps add more randomness to the system, "making the presence of any particular hidden unit unreliable" [4]. Therefore, the main motivation behind using dropout is to take a model that is sufficiently complicated which will overfit easily and perform repeated sampling and training of those samples.

## 2 METHODS

## 2.1 Network Architecture

We now describe our implemented network architecture in the following two subsections.

### 2.1.1 Encoder-Decoder Architecture

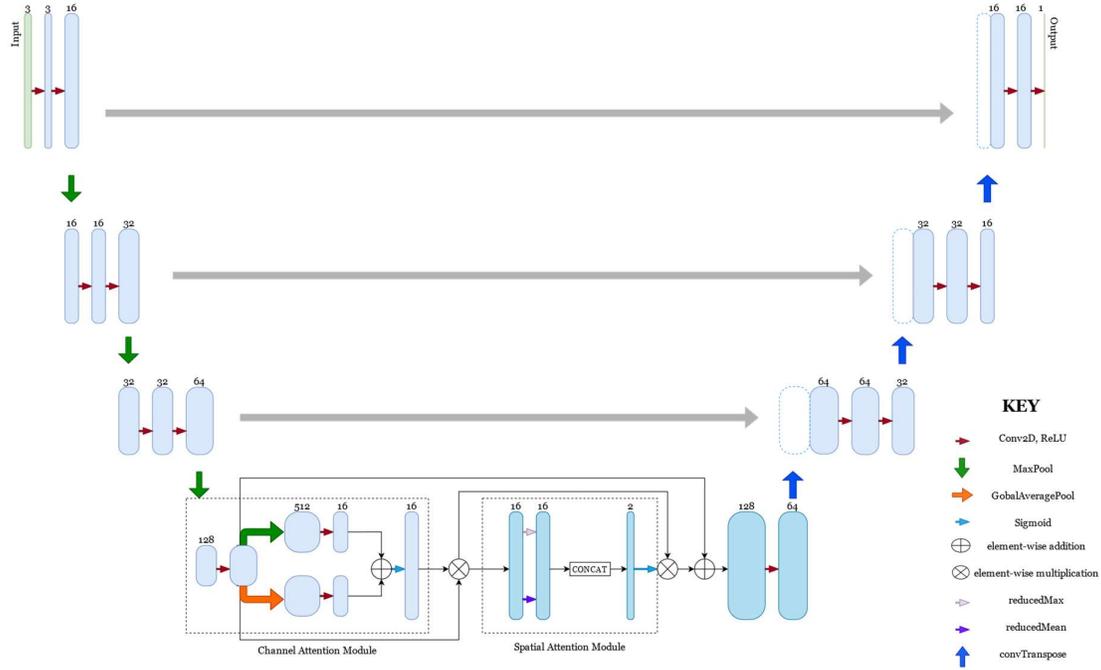

Figure 1. Model architecture diagram

The CA-UNet network architecture is shown in Figure 1. It consists of a convolutional encoding network on the left side of the model and another convolutional decoding network with an attention model present at the bottleneck. The blocks of the encoder consist of two 3x3 2D convolutions (padding of one and dilation of 1) each followed by a rectified linear unit (ReLUs) [32]. This structure is followed by a drop block of size 8x8 and a drop probability of 0.5. Then, a batch normalization [33] layer is added to regularize the network. For the downsampling of the feature vector, we apply a max-pooling operation with a kernel size of 2x2 and a stride of two. At each of the downsampling steps, we double the number of channels in the feature vector. Each decoder block has a single 2D transposed convolution that halves the number of channels in the feature vector with a kernel size of 2x2 and a stride of 2x2 followed by concatenation with the corresponding feature maps from the contracting path. Finally, at the end of the block two 3x3 convolutions, each followed by a rectified linear unit was added. The output lock of this model consists of a single 2D convolution layer with output a single channel binary segmentation map of the same resolution as the input image vector. Our output layer was followed by a sigmoid activation function to get the result.

### 2.1.2 Bottleneck Architecture

The concept of using "attention mechanisms" was first introduced in the field of Natural Language Processing (NLP) in the 2017 NeurIPS paper by Google Brain, titled "Attention Is All You Need". [23] Our convolutional attention bottleneck consists of two submodules: a Channel Attention Module and a Spatial Attention Module. Utilizing the inter-channel connection of features, a Channel Attention Module generates a channel attention map. Channel Attention focuses on 'what' is relevant given an input image since each channel of a feature map is regarded as a feature detector. We compress the spatial dimension of the input feature map to efficiently compute channel attention.

We initially use both Average-pooling and Max-pooling operations to aggregate spatial information from a feature map, resulting in two alternative spatial context descriptors: $F_{avgc}$ and $F_{maxc}$, which signify average-pooled features and max-pooled features, respectively.

After that, both descriptors are sent to a common network to create our Channel Attention Map $M_C$, where $C$ is the number of channels. A convolutional multi-Layer perceptron (MLP) with one hidden layer makes up the shared network. The hidden activation size is set to *r×1×1*, to reduce parameter overhead, where *r* is the Reduction Ratio. We use element-wise summing to integrate the output feature vectors after applying the shared network to each descriptor. Concisely, the channel attention is calculated as follows:

$$M_C(F) = \sigma\left(MLP(AvgPool(F)) + MLP(MaxPool(F))\right) \quad (1)$$

$$M_C(F) = \sigma\left(W_1\left(W_0(F_{avgc})\right) + W_1(W_0(F_{maxc}))\right) \quad (2)$$

where $\sigma$ denotes the sigmoid function. Note that the MLP weights, $W_0$ and $W_1$, are shared for both inputs and the ReLU [32] activation function is followed by $W_0$.

Spatial Attention Module uses the inter-spatial relationship of features to build a spatial attention map. We apply a convolution layer to the concatenated feature descriptor to construct the spatial attention map $M_S(F) \in RH \times W$, which encodes where to emphasize or suppress.

We use two pooling methods to aggregate channel information from a feature map, resulting in two 2D maps: $F_{avgs} \in R_1 \times H \times W$ and $F_{maxs} \in R_1 \times H \times W$. In short, the spatial attention is computed as:

$$M_S(F) = \sigma\left(f_{7\times7}([AvgPool(F); MaxPool(F)])\right) \quad (3)$$

$$M_S(F) = \sigma\left(f_{7\times7}([F_{avgs}; F_{maxs}])\right) \quad (4)$$

where, $\sigma$ denotes the sigmoid function and $f_{7\times7}$ represents a convolution operation with the filter size of *7×7*.

The input in the Attention Block is first multiplied with the Channel Attention. Then the output is passed through Spatial Attention and then multiplied by the output of the multiplied Channel Attention. The final output from the Attention is added with the input like Residual blocks. The input in the Channel Attention is passed in two parallel streams—Max Pool and Average Pool. The two parallel streams are then passed through the Conv2D-ReLU-Conv2D block parallelly. The output from the parallel blocks is then added and passed through a sigmoid. The input in the Spatial Attention is again passed in two parallel streams—maximum output as dimension *1* and average output at dimension *1*. The average and the max results are then concatenated and passed through a *7×7* Convolution. The final output is passed through a sigmoid activation function to bound the output mask pixel values within 0 to 1 .

## 2.2   Dataset

### 2.2.1   Introduction

The Cityscapes dataset [2] is made up of a variegated set of video-sequenced images from fifty cities in Europe, mostly Germany and some other cities. The resolution of each image is 2048 x 1024, with 16 bits RGB color-depths. However, for easier usage, they also provide 8-bit RGB by introducing a log-based compression curve to the images. The authors created a group of five thousand finely annotated images, with the training and validation set consisting of 3425 images, and 1525 images used for testing purposes. They also provide us with twenty thousand other coarsely annotated images for performing other training tasks if required. The pictures were captured from a moving vehicle, for several months, following spring, summer and fall seasonal differences. However, the author explicitly stated that the lack of more adverse weather conditions in the data was due to lack of more advanced resources, although their camera system is SOTA for automation purposes; something we try to make more varied by our augmentation suggestions and patterns as we will go on to mention in the paper. The annotations were described to consist of layered polygons and prepared in-house with the help of the LabelMe [24] tool.

## 2.2.2 Statistical Analysis of the Dataset

To gauge at our data set images and the masks we created, we performed some basic analysis and testing. Firstly, on the distribution of images and the cities they were collected from. We notice our training set spans across eighteen cities, our validation across three and our testing across six cities, all situated in Germany, while the number of images across both the actual image and mask count to 2975 for the training, 500 for validation, and 1525 for the test images. [2]

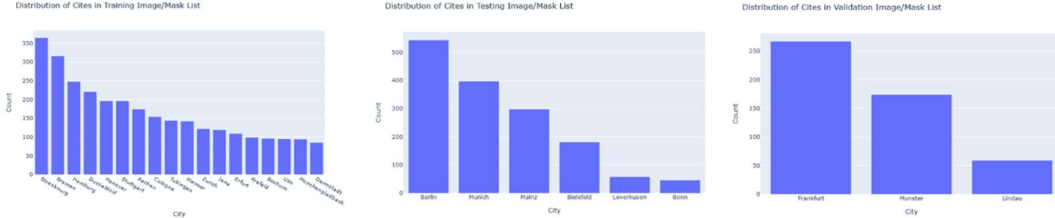

Figure 2. City-wise Distribution of the Cityscapes Dataset

We moved on to conduct several hypothesis tests to understand whether our drivable vs. non-drivable areas might be similar across our chosen training and validation data using our self-built functions to help preprocess our binary mask pixel data into drivable and non-drivable count and normalize the numeric values while considering a significance level $\alpha = 0.05$ or $5\%$.

In figure 2 we show a city-wise distribution of the images in the training, validation and testing dataset. As we see they all contain images from different cities ensuring that we do not overfit our model to the roads structure of a single city.

We first perform the Student's t-test for two samples having unequal sample size, assuming similar variance under the null hypothesis that they have identical means, where we notice our p-values $p \geq \alpha$ ($p = 0.2990$ for drivable and $p = 0.0772$ for non-drivable) in the case of both drivable as well as non-drivable areas denotes our observations are unlikely to have been misinterpreted, so we reject our null hypothesis. We have,

$$t = \frac{\bar{X} - \bar{Y}}{\sqrt{\frac{(x-1)s_X^2 + (y-1)s_Y^2}{x+y-2}} \sqrt{\frac{1}{x} - \frac{1}{y}}} \tag{5}$$

where $t$ is the t-statistic, $\bar{X}$ is the sample mean of the first sample, $\bar{Y}$ is the sample mean of the second sample, $s_X^2$ is the first sample variance, $s_Y^2$ is the second sample variance, and $x$ and $y$ are the sample counts of the first and second sample respectively.

We performed the Bartlett's test next under the null hypothesis that samples are from the population with equal variances. The values from drivable pixels show that the observed outcome would be unlikely under the null hypothesis ($p = 0.0086$) whereas in non-drivable, they become statistically significant ($p = 0.5466$). We have,

$$\chi^2 = \frac{(x+y-2)\ln\left(\frac{(x-1)s_X^2 + (y-1)s_Y^2}{x+y-2}\right) - \left((x-1)\ln(s_X^2) + (y-1)\ln(s_Y^2)\right)}{1 + \frac{1}{3}\left(\frac{1}{x-1} + \frac{1}{y-1} - \frac{1}{x+y-2}\right)} \tag{6}$$

where $\chi^2$ is the Bartlett's test statistic which approximates to a $\chi^2_{1,0.05}$ distribution, $s_X^2$ is the first sample variance, $s_Y^2$ is the second sample variance, and $x$ and $y$ are the sample counts of the first and second sample, respectively.

Next, we drew and compared the scatters for drivable vs. non-drivable areas which allows us to gauge at some overlying similarities between our test and validation masks which gives us some basic intuition on why our model performs well.

Finally, we attempted at approximating the Jaccard Index for pixel data in masks (IOU computation). We consider the Jaccard coefficient $J$ for two samples $X_i$ and $Y_i$ from sets $X$ (Training) and $Y$ (Validation) to be calculated by,

$$J(X_i, Y_i) = \frac{|X_i \cap Y_i|}{|X_i \cup Y_i|} = \frac{|X_i \cap Y_i|}{|X_i| + |Y_i| - |X_i \cap Y_i|} \qquad (7)$$

and the approximated index $J(X, Y)$ was measured by the assumption of Law of Large Numbers for the data (i.e., sample size of pairs chosen is large enough) and can be stated as,

$$J(X, Y) = \sum_{i=1}^{n} J(X_i, Y_i) = \sum_{i=1}^{n} \frac{|X_i \cap Y_i|}{|X_i| + |Y_i| - |X_i \cap Y_i|} \qquad (8)$$

We considered five hundred samples of 1 train-mask and 1 validation-mask and compute and approximate the index at each iteration. We notice a significantly large Jaccard Index $J(X, Y) = 0.8457$, which in turn helps understand why our neural network architecture learns well.

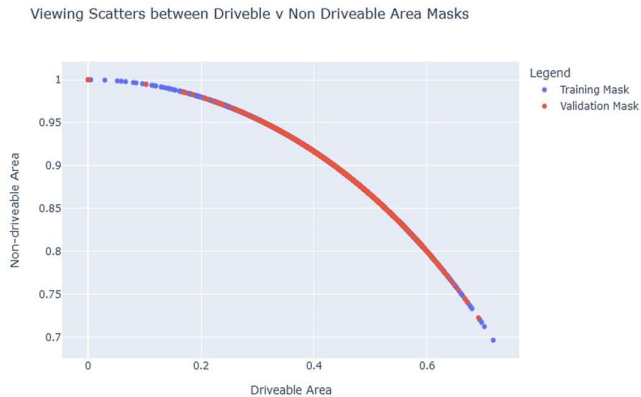

Figure 3. Correlation between Drivable and Non-Drivable Area in the Dataset

In Figure 3, we notice that when comparing the spreads of drivable and non-drivable regions, the spread of the validation masks falls mostly within the training mask, which is a good indicator how our model learns quickly, and within a certain number of epochs as we go on to state later.

## 2.3 Augmentation

Lately, Deep Learning Applications have been performing extremely well when it comes to tasks requiring segmentations. But, to prevent the long-lasting issue of overfitting of data due to lack of big datasets in multiple domains, data augmentation has become an all-round solution to tackle this. In our problem where we deal with the Cityscape dataset, we too have applied numerous geometric, affine, and pixel-level transformations to improve the overall quality of our dataset.

Firstly, pixel level transformations like Rotation Flip, Random Crop, Random Scale, and Mix-Cut have been used to bring in more randomness to our dataset, while trying to see how the model learns and performs in overall more abstract scenarios.

Next, we implemented the two famous augmentations CLAHE [35] and Random Gamma which have played a leading role in our data augmentation pipeline to remove our model's dependency on the quality and brightness of the original training images.

Adaptive Histogram Equalization (AHE) is an image processing technique used to ameliorate contrasting in digital images, differing from the ordinary Equalization methodology in that this computes several histograms in which each of them corresponds to a distinct section of the image, and uses them to redistribute the lightness values of the image. [34] It is therefore apt for "enhancing the local contrast and improving the edge definitions in each region of an image." However, AHE tends to "over-amplify noise in homogeneous regions of an image." [34]

A variant of AHE called Contrast Limited Adaptive Histogram Equalization (CLAHE) [35] prevents this by limiting the amplification. CLAHE has also been used for a long time. A good example includes a paper from 1974 for enhancing cockpit displays [25], which should also work well for the cityscape dataset.

The usage of Gamma Correction [36] matters in the sense if one has any interest in displaying an image accurately on a computer screen as it controls the overall brightness of an image. Images which are not suitably gamma corrected can look either washed out out or too dark. Trying to reproduce colors accurately also requires some knowledge of gamma. Varying the amount of gamma correction changes not only the brightness, but also the ratios of red to green to blue. [36] Random Gamma Augmentation plays a major role in increasing the variation of the dataset and considering all the possible cases, considering human perception to brightness or luminance as produced by the RGB colors, with Green being the most luminous followed by Red and finally Blue in terms of the combined proportion as seen by human eye.

Our next augmentation ColorJitter, performs similarly to CLAHE but is more random in augmenting the brightness, contrasts, and saturation of our dataset. A good usage of this implementation is shown by Simonyan et al. [15] in their 2015 paper working on large-scale image recognitions where they used the ImageNet dataset. Downscale is also implemented with a certain low probability as it plays around with image quality well.

Following up with these augmentations, we added several blurring techniques to make realistic portraits bringing in a certain clumsiness and imperfections in our datasets, in the form of random gaussian blurs (GaussianBlur), blurring produced by glass (GlassBlur), saturations in colors (HueSaturation) and blurring produced by motion (MotionBlur). These would provide our dataset with more robustness and will make it look to be as if clicked from separate locations, be it through a closed glass window or a moving bicycle, or inside a car with windows drawn up portraying a mixture of both motion and glass blurring effects.

We also added augmentations in forms of imperfections normally caused by a camera's quality, like noises (ISONoise) which helps create noise in the dataset caused by noise created by camera sensors and different distortions (OpticalDistortion) like purple fringing caused by pixel-level errors. Usage of camera errors related augmentations were inspired by the recent paper by Ouyang et al. [26] in their paper related to camera simulations caused by neural networks.

We used Posterize to randomly reduce some bits from color channels to gauge if the model can still differentiate between drivable roads and lanes on the side due to reduction of color bits.

Finally, we saw that our dataset was monotonous in terms of the weather conditions, with the entire dataset being taken in perfect weather conditions. So, we included more randomness to our dataset by adding various weather simulations of Random Fog, Random Rain, Random Snow and Random Sun flare so that we can work with a more realistic dataset giving us several different conditions and possibilities to choose from.

Figure 4 contains a visual representation of all the different types of augmentations applied in our model pipeline to increase the size of our initial dataset. We see that some of the more complex augmentations make the visual identification of the drivable road area impossible, so a machine learning approach must be taken to tackle these scenarios.

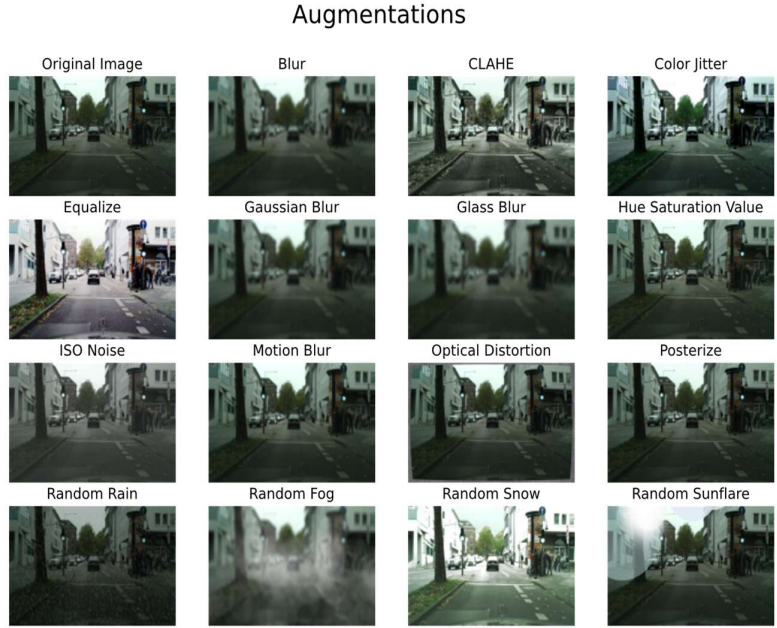

Figure 4. A visual representation of some of the augmentations applied

## 3 RESULTS

To measure the performance of the CA-UNet models and its performance with added augmentations we have tested them on the widely used cityscape dataset. For the implementation, we have used the PyTorch [37] framework. The model is trained on a single GPU machine with 32GB of RAM and Nvidia GTX 1660ti GPU.

### 3.1 Train Setting

We chose the popular Adam optimizer with a weight decay of 0.00001 and betas of 0.9 and 0.999 respectively as our optimizer along with Binary cross-entropy as our loss function. The model is trained from scratch with Kaiming Initialization. To keep the model compact and fast we sent the number of channels in the initial convolutional layer at 16. The learning rate is set at 0.001 for the first one hundred epochs, then it is reduced to 0.0001. The dropblock discard size is kept at 7 for the final model with a linear drop rate schedule of 0.05 to 0.25.

### 3.2 Evaluation Metrics

We have used the following metrics in our calculations:

**Accuracy:** A statistic that sums up how well a model performs across all classes. It is helpful when all the classes are equally important. It is computed as the ratio between the number of right guesses and the total number of forecasts.

$$Accuracy = \frac{True_{+ve} + True_{-ve}}{True_{+ve} + True_{-v} + False_{+v} + False_{-ve}} \quad (9)$$

**Jaccard Index:** The area of overlap between the resultant segmentation that we expect versus the actual truth is divided by the area of union between the predicted segmentation and the actual truth to get the Jaccard Index (also known as Intersection over Union or IOU).

$$J(X,Y) = \frac{|X \cap Y|}{|X \cup Y|} = \frac{|X \cap Y|}{|X| + |Y| - |X \cap Y|} \tag{10}$$

**Precision:** It is measured as the ratio of the number of correctly identified Positive samples to the total number of classified Positive samples (either correctly or incorrectly).

$$Precision = \frac{True_{+ve}}{True_{+ve} + False_{+ve}} \tag{11}$$

**Recall:** It is determined by dividing the total number of Positive samples by the number of Positive samples accurately categorized as Positive, measuring the model's ability to recognize Positive samples. The higher the recall, the greater the number of positive samples found.

$$Recall = \frac{True_{+ve}}{True_{+ve} + False_{-ve}} \tag{12}$$

**Dice:** It is used as a measure to gauge the similarity between two samples.

$$Dice = 2\frac{Precision \times Recall}{Precision + Recall} \tag{13}$$

**Specificity:** It is a measure of the ratio of true negatives derived from a test divided by all the negatives in the set (including the false positives).

$$Specificity = \frac{True_{-ve}}{True_{-ve} + False_{+ve}} \tag{14}$$

**Matthew's Correlation Coefficient:** Also, known as the Phi Coefficient, it is appropriate for measurements of performance in binary classification tasks for two categories having different data lengths, and can be directly derived from the values for a given confusion matrix itself.

$$MCC = \frac{(True_{+ve} \times True_{-ve}) - (False_{+ve} \times False_{-ve})}{\sqrt{(True_{+ve} + False_{+ve}) \times (True_{+ve} + False_{-ve}) \times (True_{-ve} + False_{+ve}) \times (True_{-ve} + False_{-ve})}} \tag{15}$$

## 3.3   Inference Speed Measurement

While measuring the inference speed for the model we used a single GTX 1660ti GPU and set the batch size to 1. The CUDA version we have used is 10.2 while the PyTorch version is 1.1. To reduce the problem of occasionality, we ran the same network twenty times and calculated the average inference speed under the scaled input resolution of 1024 x 512 of the Cityscapes dataset.

As shown in figure 5 the Average inference time across all the twenty model runs came up to be 0.00497 secs per frame on a GTX 1660ti which contains 1536 Cuda cores.

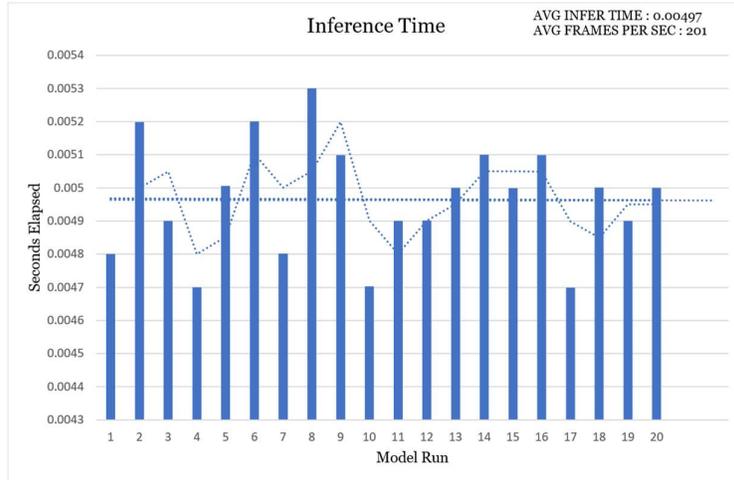

Figure 5. Average inference times vs model runs

## 3.4  Results on Cityscapes Dataset

In this part of the paper, we demonstrate the drivable area segmentation ability of our CA-UNet by comparing to other state-of-the-art models. The accuracy of CA-UNet is 98.79% on the augmented training data, and 97.02% on the validation data. As seen by the results on the validation data our model achieves a dice score of 0.9549, a f1 score of 0.97, an Jaccard Index (IoU) score of 0.914 and a MCC score of 0.9329. The most interesting point about our model is that the number of parameters in CA-UNet is much lower than other comparable CNN segmentation networks shown in table 1.

Table 1. Number of parameters in different model

| Models | Total | Trainable | Non-trainable |
| --- | --- | --- | --- |
| AG-Net [27] | 93,35,340 | 93,35,340 | 0 |
| DDRNet-23 [28] | 20100578 | 20100578 | 0 |
| SegNet [17] | 29502358 | - | - |
| 23 Layers U-Net | 21,58,705 | 21,58,705 | 0 |
| 18 Layers U-Net | 5,35,793 | 5,35,793 | 0 |
| U-Net + SA | 5,35,891 | 5,35,891 | 0 |
| SD-UNET [29] | 5,35,793 | 5,35,793 | 0 |
| SA-UNET [30] | 5,38,707 | 5,37,299 | 1408 |
| **CA-UNet (Ours)** | 4,89,588 | 4,89,588 | 0 |

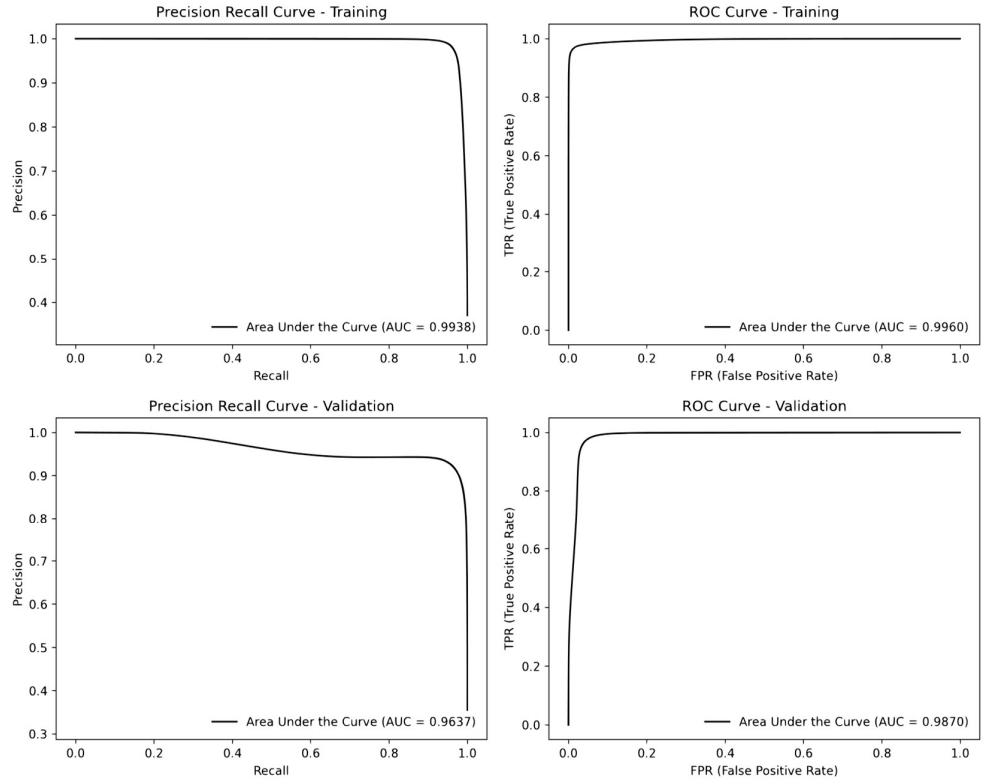

Figure 6. The PR and ROC curves on the cityscapes training and validation datasets

As shown in Figure 6, the Receiver Operating Characteristic curve (or ROC curve) for both the training data and the validation data give us high AUC (Area under the curve) scores of 0.9960 and 0.9870 respectively, indicating our model is a very capable binary classifier between drivable pixels and non-drivable pixels in an image. Furthermore, Figure 6 also shows us the Precision-Recall curves and their respective AUC scores of 0.9938 and 0.9637 which quantifies our model's ability to have a high positive predictive value while maintaining a high sensitivity.

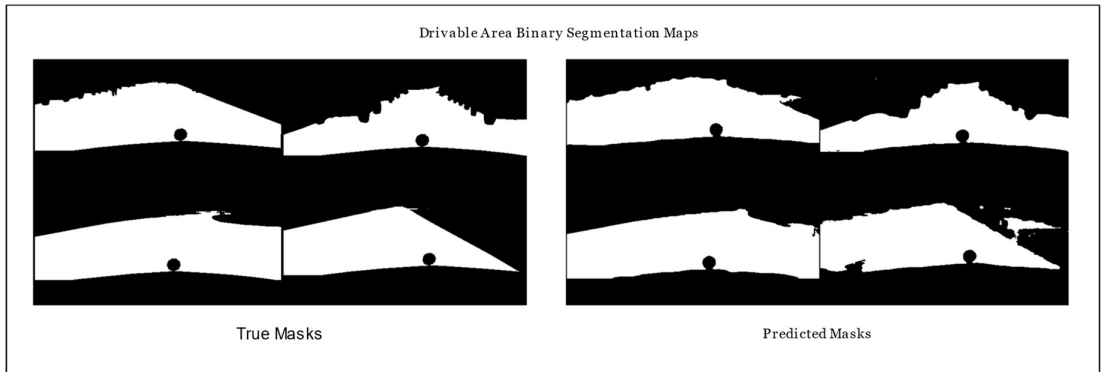

Figure 7. True and Predicted Segmentation Masks from our model

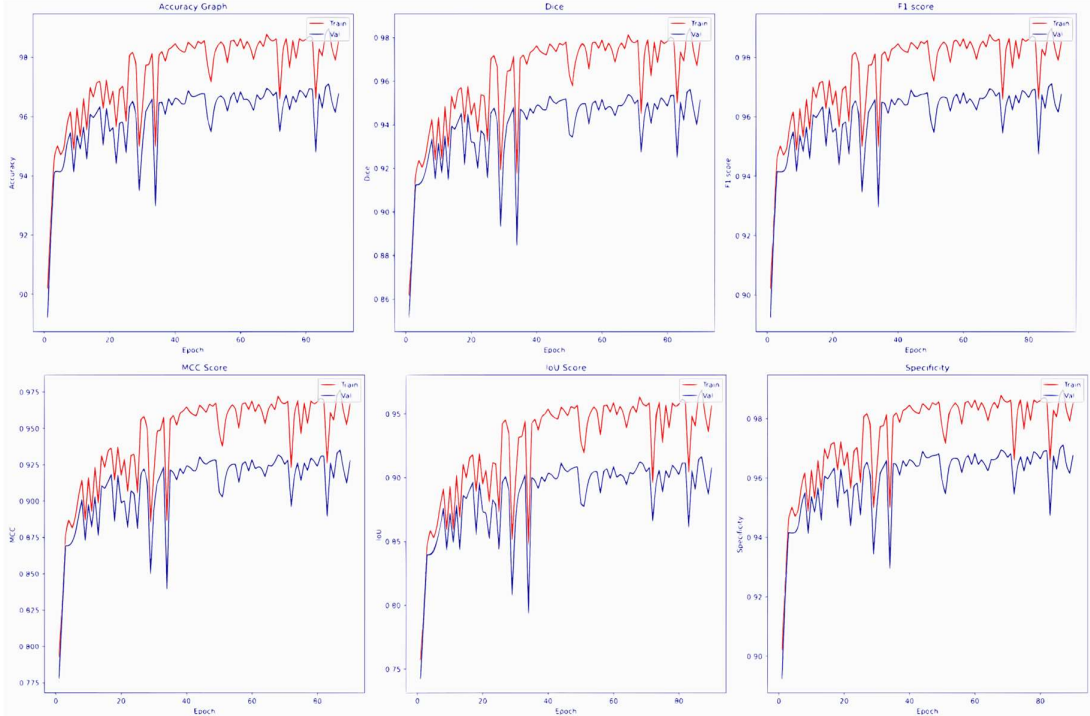
Figure 8. Performance metrics of the model trained till one hundred epochs

In Figure 7. we have shown a few examples of the segmentation masks generated by our model on the test dataset with reference to the true binary masks. We can see that our model is giving us an accurate binary map of drivable area. However, in the fourth image we can see that there are some false positives being predicted by the model. Our model has been trained for one hundred epochs and the resulting performance metrics of the model during the training process has been shown in Figure 8. Our model reaches a high specificity score of 0.98 and IoU score of 0.95 on the training data set which further stands as a proof to the ability of our model to generate accurate drivable area segmentation masks in real time.

## 3.5 Ablation Experiments

### 3.5.1 Ablative Experiment of Augmentations

Taking inspiration from Ronneberger et al [5] and Hong et al [28], we have experimented with and applied several forms of affine and non-affine image augmentations to our dataset to improve the performance of our model and make it more robust to weather and camera placement changes without compromising on the speed of the model. Drivable-area detection is only one of many processes involved in the panoptic machine vision for self-driving vehicles. Our model has been designed to run in tandem with other object detection and lane prediction systems while ensuring real time deductions on a limited computing power. Adding non-rigid transformation such as Elastic Transform and Grid Distortion modified the training image to scenes that was not realistically available during test time, so we dropped those from our augmentation pipeline. Furthermore, we performed an ablative test on our augmentation and found that the model drastically overfitted to light augmentations such as random cropping and angle rotations, within 10 epochs and gave us a accuracy of 96% on our training data set while implementing stronger augmentations such as random weather effects, motion blur and optical distortion along with the previously mentioned augmentations prevented our model from overfitting too quickly while giving us a model accuracy of 98.79%. In figure 9 and figure 10 we have provided an accuracy versus epochs trained plots for light and heavy augmentations respectively till fifty epochs on a smaller augmented subset of the total dataset.

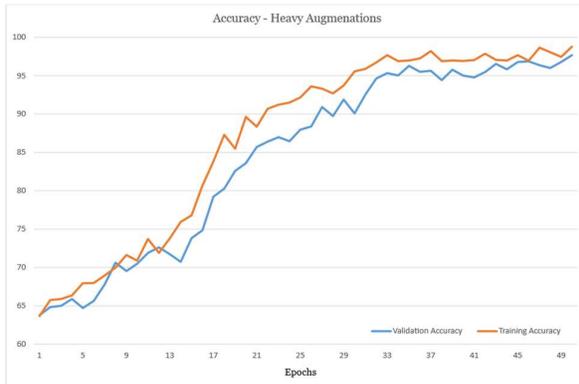
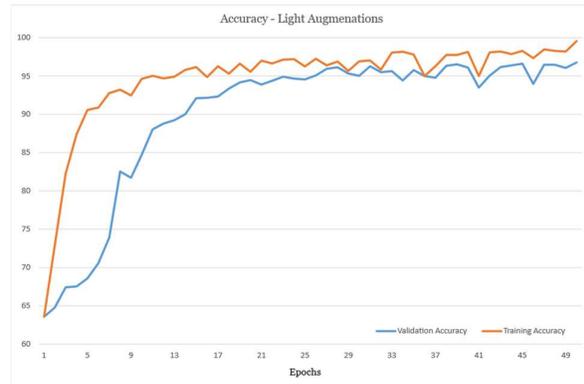

Figure 9. Plot of accuracy vs epochs trained using light augmentations

Figure 10. Plot of accuracy vs epochs trained using heavy augmentations

## 4 CONCLUSION

In this paper, an extension to the exceedingly popular U-Net architecture is proposed with the addition of convolutional and spatial attention networks to extend the feature representation capability of the model, along with the use of dropblock and batch normalization [33] for network regularization. The model has been tested rigorously on the task of drivable area detection. To reduce the overfitting of the model on the training data ambitions data augmentation processes has been applied. The evaluations done on the widely available cityscapes dataset demonstrate the effectiveness of both the attention module and the regularization techniques applied. The augmentations applied to the training data enabled the model to learn from varied training data such as obscuring weather conditions and camera instability. In future we would like to explore the scope of using different encoder-decoder structures with the same training and regularization setup.